\documentclass[11pt,a4paper]{article}
\usepackage{soul}
\usepackage[hyphens]{url}
\usepackage{hyperref}
\usepackage[hyphenbreaks]{breakurl}
\usepackage[hyperref]{acl2020}
\usepackage{times}
\usepackage{latexsym}

\usepackage{microtype}

\aclfinalcopy 

\newcommand{\lineacross}{\rule{\linewidth}{1pt}}
\newcommand{\scare}[1]{'{#1}'}
\newcommand{\lingform}[1]{\textit{#1}}
\newcommand{\err}[1]{{\color{red} \ul{#1}}}

\title{A Gold Standard Methodology for Evaluating Accuracy in Data-To-Text Systems}

\author{Craig Thomson \\
  University of Aberdeen \\
  Aberdeen \\
  UK \\
  \texttt{c.thomson@abdn.ac.uk} \\\And
  Ehud Reiter \\
  University of Aberdeen \\
  Aberdeen \\
  UK \\
  \texttt{e.reiter@abdn.ac.uk} \\}

\date{}

\begin{document}
\maketitle
\begin{abstract}
Most Natural Language Generation systems need to produce accurate texts.  We propose a methodology for high-quality human evaluation of the accuracy of generated texts, which is intended to serve as a gold-standard for accuracy evaluations of data-to-text systems.  We use our methodology to evaluate the accuracy of computer generated basketball summaries.  We then show how our gold standard evaluation can be used to validate automated metrics.
\end{abstract}

\section{Introduction}

In most contexts, it is essential that texts produced by data-to-text Natural Language Generation (NLG) systems accurately communicate input data.  Hallucination and other forms of inaccuracy are unacceptable in NLG application contexts such as journalism, financial reporting, and medical patient information.  For example, it is not acceptable to give a doctor incorrect information about a patient.  This means that it is essential that NLG developers be able to evaluate whether texts produced by their systems are accurate or not.

We propose here a methodology (protocol) for high-quality human evaluation of the accuracy of generated texts.  The methodology focuses on identifying and categorising specific accuracy errors in a text; hence it is quite different from protocols which ask people to assess the overall accuracy of a text on a scale

Existing work on detecting mistakes and hallucinations in NLG texts has largely focused on short texts which communicate relatively simple data.  For example, \citet{dusek-etal-2019-semantic} looked at slot-error-rate in the E2E challenge \citep{DUSEK2020123}, which involved generating short sentences (13 words on average) which communicated 8 attributes.  Our goal is to develop techniques which can be used to evaluate accuracy in longer texts (hundreds of words) which communicate complex data and possibly insights (eg, trends and best/worst) derived from the source data. This is more challenging task; longer texts can have contextual errors which are rare in short texts, and checking accuracy of insights derived from complex data is harder than checking whether a small number of attributes, for a single entity, are accurately communicated.

In this paper we specifically focus on finding accuracy mistakes in English language sports stories.  However, we believe the techniques could also be applied to other types of texts produced by data-to-text systems, including financial reports and  business intelligence, which are very important in commercial data-to-text applications \citep{gartner2020}.

Finding accuracy mistakes in a 300-word sports story using our methodology costs on the order of US\$30 in Mechanical Turk worker payments and Amazon fees, plus 30 minutes of experimenter time.  Workers were screened with a qualification task.  We intend our methodology to be a high quality \emph{gold standard} for accuracy evaluation, and encourage other researchers to find alternative and cheaper evaluation techniques which correlate well with the gold-standard presented here.

In other words, researchers developing metrics for measuring accuracy can compare the results of their metrics with our gold-standard accuracy evaluation, and use this comparison to \emph{validate} their metrics and assess how effective their metrics are at measuring accuracy.


\autoref{fig:example} shows example sentences from a sports story annotated by our methodology.  The text is an example, constructed from fragments from the output of different systems, with some manual adjustment to keep the example simple.  The materials used to perform the evaluation described below, as well as the small corpus of 21 accuracy-annotated sports stories has been released on GitHub\footnote{\url{https://github.com/nlgcat/evaluating\_accuracy}}.

\begin{figure}
\lineacross{}
The Memphis Grizzlies (5-\err{2}) defeated the Phoenix Suns (3 - 2) \err{Monday} 102-91 at the \err{Talking Stick Resort Arena} in Phoenix. The Grizzlies had a \err{strong} first half where they \err{out-scored} the Suns \err{59}-\err{42}. Marc Gasol scored 18 points, \err{leading} the Grizzlies.  \err{Isaiah Thomas added} 15 points, he is \err{averaging 19 points on the season so far}.  The Suns' next game will be \err{on the road} against the \err{Boston Celtics} on Friday.


\vspace{5mm}
List of errors:
\begin{itemize}
    \item \err{2}: incorrect number, should be 0.
    \item \err{Monday}: incorrect named entity, should be Wednesday.
    \item \err{Talking Stick Resort Arena}: incorrect named entity, should be US Airways Center.
    \item \err{strong}: incorrect word, the Grizzlies did not do well in the first half.
    \item \err{out-scored}: incorrect word, the Suns had a higher score in first half.
    \item \err{59}: incorrect number, should be 46.
    \item \err{42}: incorrect number, should be 52 .
    \item \err{leading}: incorrect word,  Marc Gasol did not lead the Grizzlies, Mike Conley did with 24 points.
    \item \err{Isaiah Thomas added}: context error, Thomas played for the Suns, but context here implies he played for the Grizzlies and added to their score.
    \item \err{averaging 19 points in the season so far}: Not checkable.  Data sources report performance per season and per game, not performance at a particular point in a season.
    \item \err{on the road}: incorrect word, The Suns will play at home.
    \item \err{Boston Celtics}: incorrect named entity, the Suns will play the Sacramento Kings

    
\end{itemize}
\caption{Example text with error annotations.  Each annotation includes an error type and a correction.  Annotators can add explanations where useful. Box score data for this game is available at
\url{https://www.basketball-reference.com/boxscores/201411050PHO.html} .}\label{fig:example}
\lineacross{}
\end{figure}

\section{Related Work}
\label{sec:related_work}

NLG systems can be evaluated using either automatic metrics or human evaluation \citep{celikyilmaz2020evaluation}.   Automatic metrics such as BLEU are not very meaningful in NLG \citep{reiter:cl2018}, especially when assessing accuracy \citep{reiter2009investigation}.  Even in machine translation, BLEU and related metrics are not meaningful unless the differences in metric scores is quite large, much larger than reported in most academic papers \cite{mathur-etal-2020-tangled}.

Human evaluation of NLG systems is usually done using Likert scales or ratings \citep{van-der-lee-etal-2019-best}.  In the context of evaluating accuracy, human evaluators are usually asked to assess the overall accuracy of a generated text \citep{reiter2009investigation}, or to compare two texts and say which text is overall most accurate \citep{REITER2005137,novikova-etal-2018-rankme}.

The Pyramid method \citep{nenkova-passonneau-2004-evaluating} in text summarisation is a complex technique for evaluating the quality of a summary from a content perspective.  It originally required substantial human input, but recently there have been attempts to automate PYRAMID analysis \citep{Yang:AAAI16}. However, PYRAMID focuses on checking whether expected content is present, not finding mistakes in unexpected content.

In the context of evaluating computer-generated sports stories, \citet{wiseman-etal-2017-challenges} showed sentences (not complete stories) to human subjects, and asked the subjects to count how many facts in the sentence were supported by game data and how many contradicted the game data.  These results were then compared to metrics based on information extraction techniques.  This was repeated by \citet{puduppully2019data} and extended to other domains by \citet{dhingra-etal-2019-handling}.

Another metric which semantically analyses generated text and compares this to the source data is SPICE \citep{Anderson:ECCV16}, which uses this approach to evaluate the quality of computer-generated image captions,

Accuracy-checking is also an issue in fact checking and verification.  The FEVER workshops and shared tasks \citep{thorne-etal-2018-fact,thorne-etal-2019-fever2} asked participants to develop techniques to identify factual errors in manually \scare{mutated} versions of Wikipedia articles \citep{thorne-etal-2018-fever}.

\section{Methodology}
\label{sec:methodology}

In summary, our approach is to ask multiple annotators to identify specific errors in a text, and categorise the errors into one of a small number of types.  We also ask annotators to provide corrections and optionally explanations of the error.  An example is shown in \autoref{fig:example}.
We then integrate the annotations into a single gold standard, based on majority opinion of our annotators.

The methodology described below has been refined based on results of pilot annotation exercises performed with a different group of participants.

\subsection{Real-world error vs not in the data?}

We ask annotators to mark up places where the text says things which are not true.   An alternative approach is to annotate places where the texts say things which are not in the system's input data.   These two approaches often agree but sometimes disagree. For example, consider the below sentence, in a context where there is no information in the input data about the next game.
\begin{quote}
    The Suns' next game will be on the road against the Boston Celtics. 
\end{quote}
Since this particular NLG system had no information available for the next game, the above sentence is pure hallucination.  However, the information could still be correct, by almost sheer luck.   If the sentence is factually accurate (ie, the Suns next game really was an away game against the Boston Celtics), we would not consider this to be an error.

The difference between the "in-the-data" and "real-world" approaches is most noticeable when there are facts or insights which are not present in the data but can be inferred from other data with high but not perfect confidence.  For example, suppose the data for a basketball game records whether the game is "home" or "away" for each team, but not the actual location of the game.  We can make strong guesses about location from this data; eg, a home game played by the Memphis Grizzlies will probably be in Memphis.  However, there are exceptions (eg, NBA Global Games).  In this case, stating that a home game for the Grizzlies was played in Memphis would always be considered an error under the "in-the-data" approach, but would only be considered an error under the "real-world error" approach if the game was actually played somewhere else (which is rare).

We believe that the "real-world error" approach is a better fit to what users want, so we use it.  But we realise that others have different views, and are happy to discuss this.  Its also worth noting that from a pragmatic perspective its probably easier for annotators who have domain expertise to detect real-world errors.  They do not need to check whether things they already know to be are true are present in the input data, and they can use existing resources (tools, websites, etc) which they are familiar with to find out what actually happened, without worrying about whether all the information in the resource is present in the NLG system's input data.

It is possible that the systems being evaluated used differing input data.  For example \citet{gong-etal-2019-table} include data from games beyond the one which is the focus of the summary.  This means that from a practical perspective we would need to create multiple user interfaces to present data to the annotators if we want annotators to check whether facts are in a system's input data set.

\subsection{Categories}
\label{sec:methodology:categories}
We ask our annotators to categorise errors into one of the following categories
\begin{itemize}
        \item {\em Incorrect number:}  This includes numbers which are spelled out as well as digits.
        \item {\em Incorrect named entity:} This includes people, places, organisations, and days of the week.
        \item {\em Incorrect word:} A word which is not one of the above and is incorrect.
        \item {\em Context error:} A phrase which causes an incorrect inference because of context or discourse.
        \item{\em Not checkable:} A statement which can not be checked; either the information is not available or it is too time-consuming to check.
        \item {\em Other:}  Any other type of mistake.  Annotators are asked to only use this category as a last resort.
\end{itemize}

These categories were developed based on our pilot experiments, with the key aim being to keep them simple and intuitive.  We hope that categorising errors will give NLG developers a better idea of where their system needs to be improved from an accuracy perspective.  Also, from the perspective of using our methodology as a gold standard, categories will help developers of alternative evaluation metrics to understand their effectiveness (\autoref{sec:experiment:relation-generation}).

We experimented with more linguistically meaningful categories such as {\em Incorrect referring expression}, but some of our annotators struggled to understand these categories.  The above categories capture important linguistic distinctions, and seem to be meaningful and sensible to our annotators.

The \emph{Not Checkable} category covers both
\begin{itemize}
    \item \emph{Statements which are very time-consuming to check}, such as how many points a player has scored half-way through the season. We do not want annotators to spend a large amount of time checking a single error.
    \item \emph{Statements which are impossible to check}, such as claims about what players were worried about.
\end{itemize}
We could have used separate categories, but have grouped these together because such statements are relatively rare (Section~\ref{sec:experiment:results}).  Note that whilst in principle \emph{Not Checkable} statements may not actually be errors, in practice they usually seem to be errors, at least with the neural network based systems we evaluated.  Not checkable statements tended to require data from previous games which the systems did not have access to.

The \emph{Other} category is intended to be used for accuracy errors that do not fit into any other category.  In our pilot experiments, some annotators used \emph{Other} extensively, including for errors which we believe could have been placed into a different category.  We kept the category, but explicitly asked annotators to only use it if absolutely necessary.

\subsection{Difficult Cases}\label{difficult-cases}
\label{sec:methodology:difficult-cases}
Our pilot experiments highlighted several cases where annotation was difficult.  One problem is that sometimes a text can be annotated in different ways.  For example, assume that Lou Williams scored 14 points and had 1 rebound, and Solomon Hill scored 30 points and had 6 rebounds.  In this case, the sentence \lingform{`Lou Williams scored 30 points and had 6 rebounds'} could be annotated in two ways
\begin{itemize}
    \item Lou Williams scored \err{30} points and had \err{6} rebounds. 
    \item \err{Lou Williams} scored 30 points and had 6 rebounds.
\end{itemize}
In other words, we can either annotate the numbers as incorrect (should be \lingform{14} and \lingform{1}) or the name as incorrect (should be \lingform{Solomon Hill}).   In such cases we ask annotators to choose the annotation with the fewest number of mistakes; for example the second annotation (\err{Lou Williams} should be Solomon Hill) in the above example.

Our pilot experiments also showed that it was difficult for annotators to annotate specific errors if the text included sentences or phrases which were completely nonsensical, such as  

\begin{quote}
    Markieff Morris also had a nice game off the bench, as he scored 20 points and swatted away late in the fourth quarter to give the Suns a commanding Game 1 loss to give the Suns a 118-0 record in the Eastern Conference's first playoff series with at least the Eastern Conference win in Game 5.
\end{quote}

There are so many errors in this sentence (especially since the game being described is not a playoff game) that annotators struggled to mark up specific errors.  We considered adding a {\em Nonsense} category, but decided against this, because it was difficult to define and also because we wanted to limit the number of categories.

Finally, our pilot experiments revealed cases where different annotators gave different results because they interpreted words differently.  For example, some people interpret \lingform{frontcourt} to mean 3 players (center, power forward, small forward\footnote{\url{https://www.sportsrec.com/8338357}}), while others interpret it to mean 2 players (just center and power forward\footnote{\url{https://en.wikipedia.org/wiki/Basketball_positions}}).   Because of this difference, annotators disagreed on whether the below sentence was an error or not.
\begin{quote}
The Bucks‘ frontcourt did most of the damage.
\end{quote}
We experimented with adding a glossary to resolve such issues.  However the glossary was not always effective (eg, an annotator who already knew what \lingform{frontcourt} meant might not check the glossary) and complicated the annotation exercise, so we dropped it.

We would like to revisit the above issues in our future research.

\subsection{Participants}
\label{sec:methodology:participants}

Annotating mistakes in the output of a data-to-text system requires time, attention to detail, and domain knowledge.  Annotators must be able to understand the stories, as well as any domain specific terminology.  It is not something which can be done quickly or trivially.

In large-scale annotation exercises, it may be easiest to simply hire annotators as research assistants.  In smaller-scale exercises, Mechanical Turk or other crowdsourcing platforms can be used.   We experimented with different MTurk strategies, and had the most success with the following approach:
\begin{itemize}
    \item Turkers are first asked to do a validation exercise, where they re-annotate a text which has already been annotated.  Turkers who annotate at least 70\% of the known accuracy errors are validated, and invited to participate in the main annotation task.  Those who fail to get 70\% were still paid for the validation task.
    \item Turkers are paid roughly US\$20 per hour.  In addition to ethical considerations, reasonable pay is essential for recruiting high-quality annotators.
    \item Turkers are highly rated.  We initially experienced problems with workers not spending time on our task and submitting poor (or no) annotations.  This stopped when we filtered for `Amazon masters' workers who also held US bachelors degrees.  These workers were proactive in seeking clarification and it was clear they wanted to do a good job.
\end{itemize}

Each of our participants filled in a short survey where they provided gender, age and gave an indication of their domain knowledge, eg how frequently they watch or play basketball.

Note that Turkers probably could not be used to annotate texts in specialised technical domains such as medicine.  In such cases, hiring annotators may be the best option.

We recommend asking three annotators to annotate each text; this makes the annotation process more robust.   We experimented with 5 annotators, but this did not have much impact on results, while significantly increasing costs and complexity.

\subsection{Combining Annotations}
\label{sec:methodology:combining-annotations}

The final step of the process is to combine the three individual annotations of each text into a single gold-standard annotation.   We do this ourselves, and it can be quite time-consuming, especially in cases where texts can be annotated in different ways.  During this stage, we also remove disagreements due to size of span (\autoref{sec:methodology:combining-annotations-span}) and disagreements due to one annotator not following the guidelines (\autoref{sec:methodology:combining-annotations-merged}).

\subsubsection{Size of span}
\label{sec:methodology:combining-annotations-span}

Since our protocol allowed for free annotation of text, annotators would sometimes differ in the number of words they highlighted to describe the same underlying error.  For example, one annotator could highlight `Boston Celtics' while a second may highlight `the Boston Celtics'.  In cases like this, where the entity which is being highlighted was clear, we manually adjusted for slight differences such as the inclusion of a determiner.  Another example is where one annotator highlights `on the road', while a second simply highlights `the road', or even just `road' for the location of an upcoming game.

\subsubsection{Annotator did not follow rules}
\label{sec:methodology:combining-annotations-merged}
In some cases annotators disagreed because one annotator clearly had not followed our annotation instructions.  In such cases, we used the annotation which had followed our instructions.  For example, our annotation guidelines state that an incorrect day-of-week should be regarded as a Name error.  Hence if annotators disagreed because some had annotated an incorrect day-of-week as a Name error but others had annotated it as a Word error, we recorded this as a Name error, since this is what our guidelines state.

Another common example was when annotators marked a construct such as `30-20' (eg, \lingform{`Boston won the rebounding battle 30-20}) as a single error even if both numbers were incorrect.  Our guidelines state this should be treated as two errors if both numbers are wrong.  So again if some annotators marked the above as one error but others marked it as two errors, we recorded this as two errors, since this is what our guidelines state.

As discussed in \autoref{sec:future_work}, these problems would be reduced or eliminated if we had a good tool and user interface for doing the annotation.

\subsubsection{Text can be annotated in different ways}
\label{sec:methodology:combining-annotations-complex}

There were also cases where annotators  disagreed with each other in ways which were not resolved by our annotation guidelines.  In such cases, we recorded the majority opinion.

Some of these disagreements could have been resolved by more detailed annotation instructions (some examples are given in \autoref{sec:experiment:improvements-to-annotation-scheme}).  However others are more fundamental.  For example, sometimes there are different ways of marking up a sentence (with the same number of errors).  For instance \lingform{`Joel Embiid led the Heat with 30 points'} could be changed to either of the true sentences \lingform{`Joel Embiid led the 76ers with 30 points'} or \lingform{`Josh Richardson led the Heat with 30 points'}.  Both methods only involve one change.  There are also more complex cases where multiple mistakes could be corrected in multiple ways.
  
Ideally the annotators would discuss such cases as a group and come to a consensus.  However this is difficult to do with Mechanical Turk.

\section{Experiment}
\label{sec:experiment}
We used our methodology to annotate a small corpus of sports stories.

\subsection{Data}
\label{sec:experiment:data}
We worked with descriptions of basketball games which were produced by neural NLG systems from box score and other game data.   We obtained 21 such stories, 7 each from the systems of \citet{wiseman-etal-2017-challenges}, \citet{puduppully2019data} and \citet{Rebuffel:IR2020}.
The stories were 289 words long on average, with the shortest being 184 words long and the longest being 452 words long.

\subsection{Annotations}
\label{sec:experiment:annotations}
We annotated all of the stories ourselves, to ensure that the process worked and that the Turkers were finding a similar number of errors to us; the annotation took us 20-30 minutes per story.

For the main exercise, we asked three Mechanical Turkers to annotate the 21 stories, using the procedure described above.   Turkers were first asked to do a validation exercise. A total of 18 workers attempted this exercise.  10 made no effort, 4 tried but did not pass, one Turker passed but did not accept subsequent work, and three Turkers passed and agreed to do the main annotation exercise.    We asked each of these three Turkers to annotate all 21 stories.  We paid the workers US\$8 per story (plus 20\% Amazon fees); this is equivalent to approximately US\$20 per hour, on the assumption that it takes 25 minutes on average to annotate a story.  All of the participants who passed our qualifying task and agreed to do the main task for us were male, aged 30-60, and played or watched basketball regularly.

Annotation was done by marking up a Word document (see supplementary data), not through a graphical user interface.    Annotation probably would be faster with a custom tool and user-interface.

\begin{table*}[!h]
    \lineacross{}
    \centering
    \begin{tabular}{l|r|r|r|r|r|r|r|r|r|r}
        error & total & all & number & name & word & context & not & other & no & no \\
        type  &       & agree &      &      &      &         & check &  & type & error \\ \hline
        number & 184 & 124 & - & 0 & 12 & 1 & 5 & 0 & 0 & 42 \\
        name & 105 & 75 & 0 & - & 4 & 2 & 0 & 0 & 3 & 21 \\
        word & 80 & 29 & 14 & 3 & - & 3 & 1 & 0 & 3 & 27 \\
        context & 19 & 7 & 0 & 2 & 1 & - & 0 & 0 & 0 & 9 \\
        not checkable & 6 & 1 & 3 & 0 & 0 & 0 & - & 0 & 0 & 2 \\
        other & 3 & 1 & 0 & 0 & 0 & 0 & 0 & - & 0 & 2 \\
        split & 21 & 0 & 12 & 5 & 16 & 9 & 4 & 3 & 0 & 14 \\
        no label & 0 & 0 & 0 & 0 & 0 & 0 & 0 & 0 & 0 & 0 \\
        no error & 0 & 0 & 0 & 0 & 0 & 0 & 0 & 0 & 0 & 0 \\

    \end{tabular}
    \caption{Confusion Matrix for Accuracy Annotations.  Table shows disagreements between annotators on error category, after annotations were corrected as described in \autoref{sec:methodology:combining-annotations-merged}. \emph{Error type} is the majority annotation or \emph{no majority} if all  annotators suggested a different annotation.  \emph{All agree} is the number of times, for this error type, where all annotators agreed.   \emph{Number, name, word, context, not checkable} and \emph{other} is the number of times one annotator choose this category when it was not the majority annotation.  \emph{No type} means the minority annotator marked this as an error but did not give a type.  \emph{No error} means the minority annotator did not annotate this as an error (either because they did not regard it as an error or because he missed it).}
    \label{tab:confusion}
    \lineacross{}
\end{table*}
\begin{table*}[!h]
    \centering
    \begin{tabular}{l|r|r|r|r|r|r|r|r|r|r}
        system & total & number & name & word & context & not checkable & other \\ \hline
        \citet{wiseman-etal-2017-challenges} & 20.3 & 9.3 & 5.1 & 5.0 & 0.4 & 0.3 & 0.1\\
        \citet{puduppully2019data} & 20.9 & 10.9 & 5.3 & 4.0 & 0.7 & 0 & 0\\ 
        \citet{Rebuffel:IR2020} & 15.0 & 6.0 & 4.0 & 2.6 & 1.6 & 0.6 & 0.3\\
        \hline
        
    \end{tabular}
        \caption{Average story error rate per system.  Note that a system which was used in real contexts would probably need a total error rate of less than one error per story in order to be usable.}
    \label{tab:bysystem}
    \lineacross{}
\end{table*}

\subsection{Results}
\label{sec:experiment:results}
We show here only the results from the Turkers.  As mentioned above, we also annotated the stories ourselves; since our annotations were quite similar to the Turkers and researcher-annotations are not part of the methodology, we will not further discuss these annotations here but will release them with the worker annotations.

We found a total of 418 accuracy errors in the 21 stories (ie, on average approximately 20 accuracy errors per story).  The breakdown by category was:
\begin{itemize}
    \item 184 number errors
    \item 105 name errors
    \item 80 word errors
    \item 19 context errors
    \item 6 not-checkable errors
    \item 3 other errors
    \item 21 errors where there was no majority annotation
\end{itemize}

\subsection{Disagreements between Annotators}
\label{sec:experiment:disagreements-between-annotators}

In \autoref{tab:confusion}, we give a confusion matrix which shows cases where where the majority of annotators thought there was an error, but the annotators did not agree on the type. The table shows disagreement after we have corrected annotations that did not follow our guidelines, as described in \autoref{sec:methodology:combining-annotations-merged}.
The biggest source of disagreements was when a minority annotator did not mark something as an error; inspection of these cases suggests that in most such instances there was an error, but the annotator missed it.

The Fleiss' kappa figure for inter-annotator agreement on error type classification was 0.79.

\subsection{Improvements to Annotation Scheme}
\label{sec:experiment:improvements-to-annotation-scheme}

Looking at specific examples of disagreement suggests several ways in which the annotation scheme could be improved and clarified. For example
\begin{itemize}
    \item \emph{One error or two}: In some cases annotators disagreed about whether two errors were present or just one. For example, if a player who was on the starting team but did poorly was described as \lingform{led the bench}, some annotators marked this as two errors (player was (A) not a leader and (B) not \lingform{on the bench} since he was a starter), while other annotators marked this as one error.
    \item \emph{Pronouns}: If a phrase contains a Name error and the following phrase contains a pronoun, is it an error if the pronoun refers to the corrected Name?  For example, in \lingform{the Bucks led for the majority of the game, as they led by double-digits}, Bucks is a Name error and should be replaced by Lakers (other team).  The pronoun \lingform{they} should also refer to the Lakers, is this a second error?
\end{itemize}

Of course, we need to try to keep the annotation protocol as simple as possible when making the above changes.

\subsection{Results by System}
\label{sec:experiment:results-by-system}
\label{sec:results_by_system}

The goal of our experiment was not to compare systems, but nonetheless we show in \autoref{tab:bysystem} the average number of accuracy errors in each system.  Please keep in mind that we obtained generated stories for different games for each of the three systems (in order to get a more varied set of texts in our experiments).  If our goal was to compare the systems, we would have generated stories for the same game for each system.

With these caveats in mind, \autoref{tab:bysystem} suggests that \citet{puduppully2019data} and \citet{wiseman-etal-2017-challenges} have similar profiles for accuracy errors.  \citet{Rebuffel:IR2020} has fewer errors overall and fewer number errors, but more context errors.  Again we would need to redo the experiment with a different design (getting each system to generate stories for the same games) in order to properly compare the systems, but this does show how our methodology can at least in principle be used to evaluate and compare systems.


Incidentally, the number of accuracy errors made by all three systems is far higher than would be acceptable in a published sports story.  The acceptable error rate depends on venue and context, but it almost certainly would need to be less than one error per story, with 0.1 error per story being a better target.  Of course sports stories are primarily entertainment; data-to-text systems which generate medical reports which support clinical decision making would probably need an error rate of less than 0.001 per story in order to be useful.

\begin{table*}
\lineacross{}
    \centering
    \begin{tabular}{l|c|c|c|c|c|c}
        measurement & number & name & word & context & not checkable & other \\ \hline
        Recall & 0.343 & 0.388 & --- & --- & --- & --- \\
        Precision & 0.571 & 0.755 & --- & --- & --- & --- \\
        
        \hline
        
    \end{tabular}
        \caption{Recall and Precision of RG metric per error type.  --- indicates that RG reported no errors of this type.  In fact, RG is not capable of reporting such errors as all of the data types from it\'s tuples are either names or numbers.  We accept as correct recall, any instance where RG identifies an error.}
    \label{tab:rg_recall_precision}
    \lineacross{}
\end{table*}

\subsection{Validating a Metric}
\label{sec:experiment:relation-generation}

We hope that our protocol can be used to {\em validate} metrics, that is to see how effective metrics are at measuring accuracy. Accordingly, we used our results to measure the effectiveness of the 
\emph{Relation Generation} (RG) metric, which uses information extraction and was proposed by \citet{wiseman-etal-2017-challenges} as a way of measuring the accuracy of generated sports summaries.  In fact Wiseman et. al argued that `post-hoc information extraction is significantly easier than generation itself' and this should be leveraged to optimise systems.

RG uses an information extraction model trained to link facts in the text to tuples in the data.  It operates on the same training corpus as the language generation model itself.  During evaluation, RG extracts tuples from generated text, then compares these with known tuples in the data.  Each tuple has a data type, which can be matched to the error types we define in \autoref{sec:methodology:categories}.  RG only reports errors of the `name' and `number' categories; it cannot detect word, context, not-checkable, and other errors.

We used an extended version of the RG annotation algorithm \citep{INLG2020-AddingData} which detects additional relations such as the day of the week each game was played on and subsequent opponents for each team.  We trained IE models following the general procedure proposed by \citep{wiseman-etal-2017-challenges}.  We trained with 3 random seeds and 5 learning rates, then manually chose the best 3 LSTM and best 3 Convolutional models to ensemble.

We show recall and precision in \autoref{tab:rg_recall_precision}.  Note that recall in this context is not that of the IE models themselves, but rather the fraction of accuracy errors in our gold standard annotation which were detected by RG. Precision is the fraction of errors reported by RG which were in our gold standard.

As with \autoref{sec:results_by_system} our goal is to illustrate how our evaluation protocol can be used to measure the efficacy of the metric itself, to enable it's use as a proxy for the gold standard.  Our results suggest that in addition to its inability to detect word, context, not-checkable, and other errors, RG also misses many name and number errors.

As described in \autoref{sec:methodology:difficult-cases}, when annotators were faced with multiple ways of marking up a text, we asked them to choose the one with the least number of errors.  The RG metric in such cases usually reported many number errors, rather than a single name error.  When comparing RG with our gold standard we manually adjusted for this, and considered such cases to be a match when computing recall and precision.

Systems evaluated in previous works using the RG metric have reported values in excess of 90\% \citep{Rebuffel:IR2020}.  However, if the metric itself only recalls 35-40\% of gold errors it may not be a reliable measure.

RG relies upon the data and the text presenting facts in the same form.  Hence it can capture things like `Solomon Hill scored 30 points', but would struggle with `Solomon Hill posted a double-double', an aggregate statistic where the player must have double digit counts in exactly two of points, rebounds, assists, blocks or steals\footnote{\url{https://en.wikipedia.org/wiki/Double-double}}.  This is common basketball terminology which the systems we investigated almost always used incorrectly.

Another common mistake for RG involved sentences such as:
\begin{quote}
The Raptors got off to a hot start in this game, out-scoring the Heat 64-52 in the first half alone.
\end{quote}
The input data for the systems we tested had only a per-quarter breakdown of points, as well as a total for the whole game.  It does not provide a per-half score.  Statements made in the generated texts about points-per-half are often hallucination because of this. The RG metric was unable to detect the above example.

There are many other cases where the facts in the generated text are not of the same form present in the data, such as:
\begin{quote}
He's now combined for 37 points in the last two games, as he continues to stay very consistent for the Heat.
\end{quote}
This is an error, the actual score for this player (Hassan Whiteside) was 33 points over the last 2 games.  RG does not have access to data for the previous game and therefore cannot detect such an error.  The NLG system also does not have access to this data, yet often hallucinates such statements.

Semantic control techniques for bringing the data into closer alignment with the text \citep{dusek-etal-2019-semantic} have been shown to be useful with less complex datasets such as the E2E Challenge \citep{dusek-etal-2018-findings}.  However, aligning data and text with the level of complexity shown in the sports summaries is a more difficult task.  \citet{wang-2019-revisiting} aims to prevent generation of sentences not grounded in the data, which does bring the data and generated text into closer alignment, although at the cost of limiting the types of sentence the system is capable of generating.

We hope that comparison with gold standard annotations will lead to improved versions of RG and other metrics, as well as a better understanding of where they succeed and where they fail.  If RG could reliably detect name and number errors, but not other categories of error, it would still be useful, provided that this limitation was clearly specified and understood.

\section{Future Work}
\label{sec:future_work}
We would like to improve the annotation scheme, and ideally also create a proper annotation tool.  In addition to speeding up annotation, such a tool could encourage annotators to follow guidelines such as annotating incorrect days-of-week as Name errors.  We would also like to annotate human authored texts using our methodology in order to provide a topline for NLG systems.

We are also planning to run a shared task for accuracy evaluation, where researchers propose faster and cheaper ways of finding accuracy mistakes (either via a human protocol or with a metric), and these techniques are evaluated by seeing how closely they correlate with the gold-standard accuracy evaluation described in this paper.   The shared task is described in a companion paper \citep{INLG20sharedtask}.

\section{Conclusion}
\label{sec:conclusion}
Texts generated by NLG systems need to be accurate, but current neural NLG systems often generate texts with many mistakes (\autoref{tab:bysystem}).  In order to fix this problem, we need to be able to measure how accurate texts are.   The methodology we present here will allow developers of data-to-text NLG systems to measure the accuracy of texts produced by their systems,  It will also make it easier for other researchers to develop cheaper and quicker techniques for measuring accuracy, by giving them a gold standard which they can use to validate their ideas.

\section{Acknowledgements}
\label{sec:acknowledgments}
Many thanks to the Mechanical Turk annotators who participated in our experiment, and also to David Reiter, Tim Daniels, Rodrigo de Oliveira, and Andrew Smith for serving as pilot annotators when we were developing the methodology described in this paper.  We would also like to thank Moray Greig for being our basketball domain expert during development.  We are also  grateful for the very helpful comments on this paper from the anonymous reviewers, the Aberdeen CLAN group, David Howcroft, Clément Rebuffel, and Chris van der Lee.

We would also like to thank Sam Wiseman, Ratish Puduppully, and Clément Rebuffel for providing the generated texts from their respective systems.

The work presented here is partially funded by the Engineering and Physical Sciences Research Council (EPSRC), which funds Craig Thomson under a National Productivity Investment Fund Doctoral Studentship (EP/R512412/1).

\bibliography{accuracyeval}
\bibliographystyle{acl_natbib}

\end{document}